\documentclass{altum-style}
\usepackage[utf8]{inputenc}
\usepackage[english]{babel}
\usepackage{float}
\usepackage{natbib}

\title{Multivariate Temporal Autoencoder for Predictive Reconstruction of Deep Sequences}
\author{Jakob Aungiers}
\affil{Altum Intelligence}
\affil{\textit{research@altumintelligence.com}}

\begin{document}

\abstract{Time series sequence prediction and modelling has proven to be a challenging endeavor in real world datasets. Two key issues are the multi-dimensionality of data and the interaction of independent dimensions forming a latent output signal, as well as the representation of multi-dimensional temporal data inside of a predictive model. This paper proposes a multi-branch deep neural network approach to tackling the aforementioned problems by modelling a latent state vector representation of data windows through the use of a recurrent autoencoder branch and subsequently feeding the trained latent vector representation into a predictor branch of the model. This model is henceforth referred to as Multivariate Temporal Autoencoder (MvTAe). The framework in this paper utilizes a synthetic multivariate temporal dataset which contains dimensions that combine to create a hidden output target.}

\keywords{temporal autoencoder, deep neural network, time series, multivariate model, feature engineering, signal processing}

\maketitle
\thispagestyle{thefirstpage}

\section{Introduction}

Temporal sequence prediction problems have been studied for centuries using ever more complex methods with the aim of capturing hidden patterns within and predicting those patterns going forward. Any temporal process has drivers which determine its behavior, in theory any and all of these drivers can be modelled given enough data about that process at a point in time and a complex enough model – in practice however this is currently unfeasible for a variety of reasons, the main of which are capturing the data, computing the captured dimensionality of the data and modelling the complex interaction of many dimensions interacting in various correlated ways.

An example of the complexity of such a problem might be the seemingly stochastic path of a raindrop down a window. By all respects this raindrop would appear to be taking a random walk down the windowpane, with the left and right movements seemingly unable to be determined or modelled. Consider however having the position of every water molecule, every glass molecule, their respective temperatures and their historical interactions graph with every other molecule available as data at every granular point in time. Given this information, it is reasonable to assume that there exists a model which can be created that is accurately able to specify where the raindrop will go next, and by extrapolating, where it will end up when it reaches the bottom of the windowpane.

The problem of course with the above example is that there currently exists no such method of capturing every observable aspect of a universe at a point in time. Hence for now the best we can do is look to create a model to approximate the hidden drivers of the raindrop given the best data we can gather.

Whilst this isn’t optimal for the example raindrop problem, the good news is that there are ample problems where a large amount of data can be gathered at very fine points in time and hence a model can be created to forecast the problem process.

Processes which have a small, closed universe of potential drivers that influence their behavior are easier to forecast for greater sequential steps ahead, whereas processes which are exposed to a great variety of influencing drivers succumb to the exponential decay of accuracy through chaos and as such are only able to be modelled very short sequential steps ahead. The more influencing drivers of a system can be worked into the model however, the more accurate the prediction process will be going forward.

This research focuses on building a model which can process multivariate temporal sequences of data, which in real-world data problems act as the influencing drivers of a process and which learns to build a hyperdimensional approximate representation of the drivers and process in an unsupervised manner. This trained hyperdimensional hidden representation then acts to train a secondary predictive model branch to forecast sequential steps ahead. The model is created using a multi-branch deep neural network approach utilizing the autoencoding principle and building on a sequence to sequence approach created by \cite{sutskever2014sequence} for creating the hyperdimensional hidden state representation. The model is henceforth referred to as Multivariate Temporal Autoencoder (MvTAe).

The dataset used in this research is created to be of a toy-dataset nature used to demonstrate the MvTAe model in simple yet fully functional circumstances. This research is not concerned with the other major challenge of real-world usage concerning observation, measurement and data processing.

\section{Synthetic Multivariate Temporal Dataset}

To train and test our multivariate temporal autoencoder model we create a synthetic toy-dataset which contains several specific dimensions:

\begin{itemize}
	\item \textbf{sine\_1} : a sinusoidal wave with a cycle period of 100 timesteps and an amplitude of 1.
	\item \textbf{sine\_2} : a sinusoidal wave with a cycle period of 1000 timesteps and an amplitude of 5.
	\item \textbf{noise} : a gaussian distribution of stochastic noise between -1 and +1.
	\item \textbf{combined\_signal} : a sum of sine\_1 and sine\_2. This will be used as the Y target variable we are looking to predict and will NOT be included in the X training data that the autoencoder branch of MvTAe sees.
\end{itemize}

The dataset is created in this way as to provide a way to test our autoencoder model for several important attributes. The first sinusoidal wave is a repeating pattern over time which will test the ability of our model to capture the sequential process of this pattern. The second sinusoidal wave creates a longer term cyclical sequence pattern which our model will not be able to see in full for each training example and hence it tests the models ability to capture cyclical trends. The noise dimension adds an extra dimension of redundant information to test the models ability to identify and disregard dimensions which do not contribute to the latent drivers of the data. Finally, the combined signal will test the ability of the predictive branch of MvTAe to combine signals from the two visible dimensions into this third hidden target dimension.

\begin{figure}[t]
  \includegraphics[width=\linewidth]{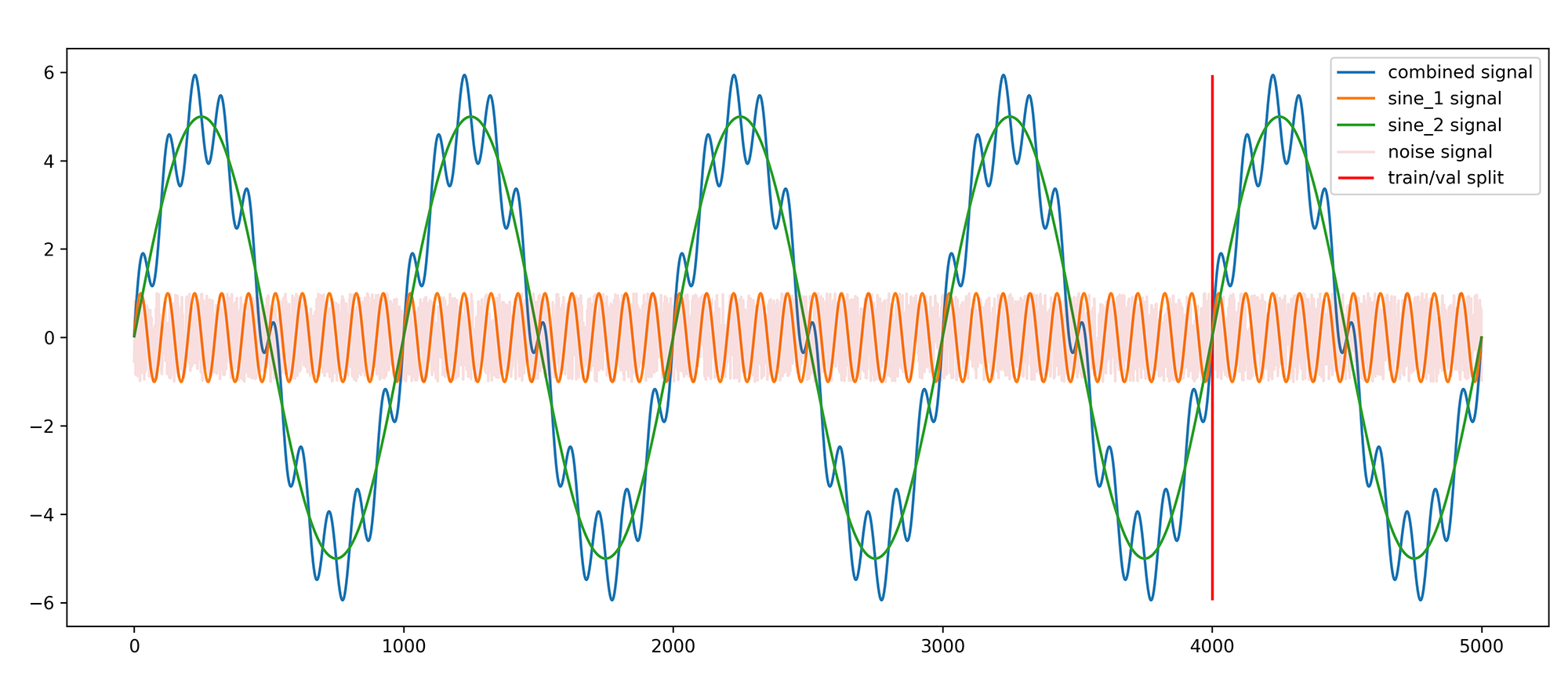}
  \caption{Synthetic multivariate temporal data across all dimensions}
  \label{fig:datasetfull}
\end{figure}

In the autoencoder branch of the model this combined signal dimension is not used as input, since in this stage the aim is to create a latent vector representation of the visible X dimensions of the dataset. In the second-stage predictive branch the combined signal is used as the Y target for future predictions.

To feed our model, the dataset is split into sliding windows of length N with step S between each window. This approach allows the training of our autoencoder branch to lookback across N temporal steps to determine relationship patterns within the temporal sequence. The Y targets of our first-stage autoencoder branch are the inverse of our inputs along the temporal axis. The Y targets of our second-stage predictive branch will be the $t_{i+1}$ combined\_signal dimension for each window of $t_{i-N} \rightarrow t_{i}$.

\begin{equation}
\label{eqn:norm}
W_{normalized} = \frac{W_{(i-N)\rightarrow i}^{k} - min(W_{(i-N) \rightarrow i}^{k})}{max(W_{(i-N) \rightarrow i}^{k}) - min(W_{(i-N) \rightarrow i}^{k})}
\end{equation}

As is standard practice when training deep neural networks for optimal converging performance \cite{WANG202082}, we normalize our data. As we are dealing with temporal data windows along multiple dimensions, we treat each window and each dimension within the window as independent in terms of normalization. What this means is that for each window W of dimension k we normalize the data independently of all other k dimensions within that window. For the normalization process itself we use standard MinMax Normalization. As such, the normalization process can be summed up as per eqn. (\ref{eqn:norm}).

\begin{equation}
\label{eqn:denorm}
W_{denormalized} = W_{(i-N) \rightarrow i}^{k} \times (hi_{i}^{k} - lo_{i}^{k}) + lo_{i}^{k}
\end{equation}

Furthermore, when used in real-world predictive applications it is usually advantageous for the final predictive output of the model to be on the absolute scale of the input data. As such, a de-normalization process is required to bring data back to the input scale. With MinMax normalization we normalize data using the min (lo) and max (hi) values of the data window and hence these values created during the normalization process are required for the de-normalization process. We define this de-normalization process as per eqn. (\ref{eqn:denorm}).

\begin{figure*}[t]
  \includegraphics[width=\linewidth]{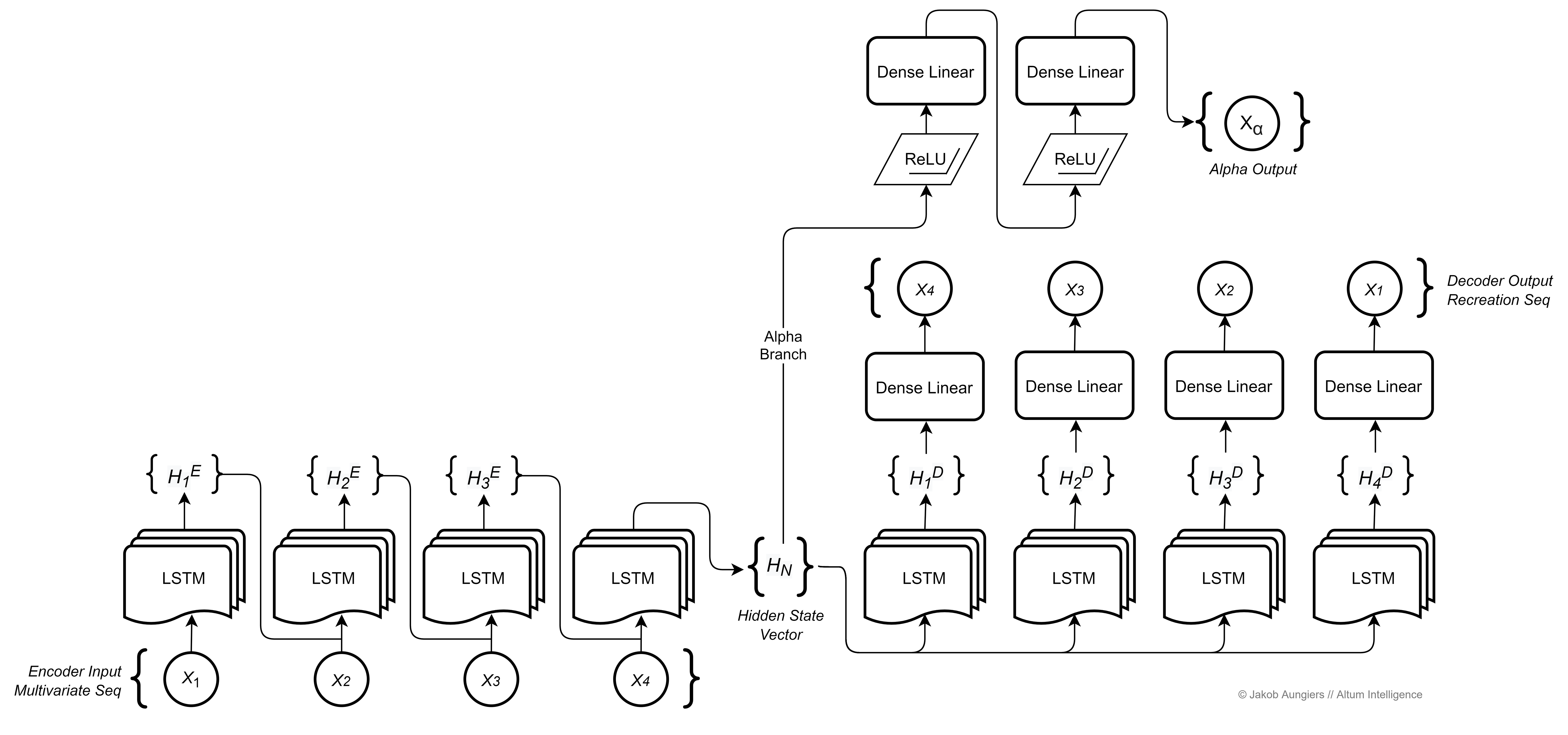}
  \caption{High-level architecture diagram of the MvTAe model}
  \label{fig:mvtaearchitecture}
\end{figure*}

\section{Multivariate Temporal Autoencoder Model (MvTAe)}

The first-stage in our predictive problem is the representation of our multidimensional temporal sequences in an optimized vector format representing the features of the multivariate series in such a way that the full series dynamics can be captured. This process can more commonly be known as feature engineering \cite{LI2020101494} and is usually a step that requires domain knowledge and a manual feature creation process when building approximations of latent drivers.

The MvTAe model acts to compress the sequence into a hidden state vector representation in an unsupervised manner, intrinsically finding latent features within the series and representing them within this state vector.

The composition of the MvTAe model is similar to that of a Sequence-2-Sequence model \cite{sutskever2014sequence} on the first branch. Several key differences however allow the MvTAe model to work more optimally for multivariate time series sequences.

The first branch of the MvTAe model is composed of two parts: an encoder which transforms the input sequence into the hidden state vector and a decoder which takes a hidden state vector and transforms it back into the original sequence, albeit in reverse. We call this branch of the model the EncoderDecoder branch.

The encoder portion takes as its input a tensor representing the multidimensional window sequence of the normalized data. This tensor serves as the input to a Long Short-Term Memory (LSTM) cells layer. The LSTM cells here take the dimensionality of the input sequence as the input dimensionality and for each sequential step return a context vector of fixed specified dimensionality. The context vector of the final sequence step LSTM cell is taken and labelled as our hidden state vector. This hidden state vector, when the EncoderDecoder is properly trained, can be regarded as a high dimensional approximation of the drivers that make up the full dimensions of the entire input sequence – in essence this is the feature vector that traditional feature engineering aims to create and which is then used with the second branch of the model to predict future sequence steps, however the creation of this feature vector/hidden state vector is done in an unsupervised way by the decoder.

The decoder structure is similar to the encoder in the sense that it is composed of the same layer of LSTM cells equal to the sequence length. The input to each of these cells is the hidden state vector created from the final context vector of the last encoder LSTM cell, copied across into each decoder cell. Note that although the hidden state vector is a LSTM contextual output, we do not treat it as a contextual input to the decoder LSTM cells, instead it is treated as a regular input and the initial cell contexts of the decoder are initialized stochastically \cite{TORRES2020113270} \cite{sordoni-etal-2015-neural}.

The decoder structure also contains an addition linear fully connected neural layer between the LSTM cell outputs and the final output. This fully connected layer enables the backpropagation training process to capture higher dimensionality linear functions within the data and hence allows the LSTM decoder layer to focus on capturing the non-linear sequential functions within the data.

The decoder output – and what makes this process unsupervised – is the same input as to the encoder, hence the model acts in an autoencoder fashion mapping $X \rightarrow X$. However one thing to note is that the decoder output targets are the reversed input of X ($\hat{X}$) hence $X \rightarrow \hat{X}$. This is done as \cite{sutskever2014sequence} found reversing the decoder targets significantly improves modelling accuracy, likely due to a higher influence of short-term dependencies within the sequences as opposed to longer term patterns.

\begin{figure}[t]
  \includegraphics[width=\linewidth]{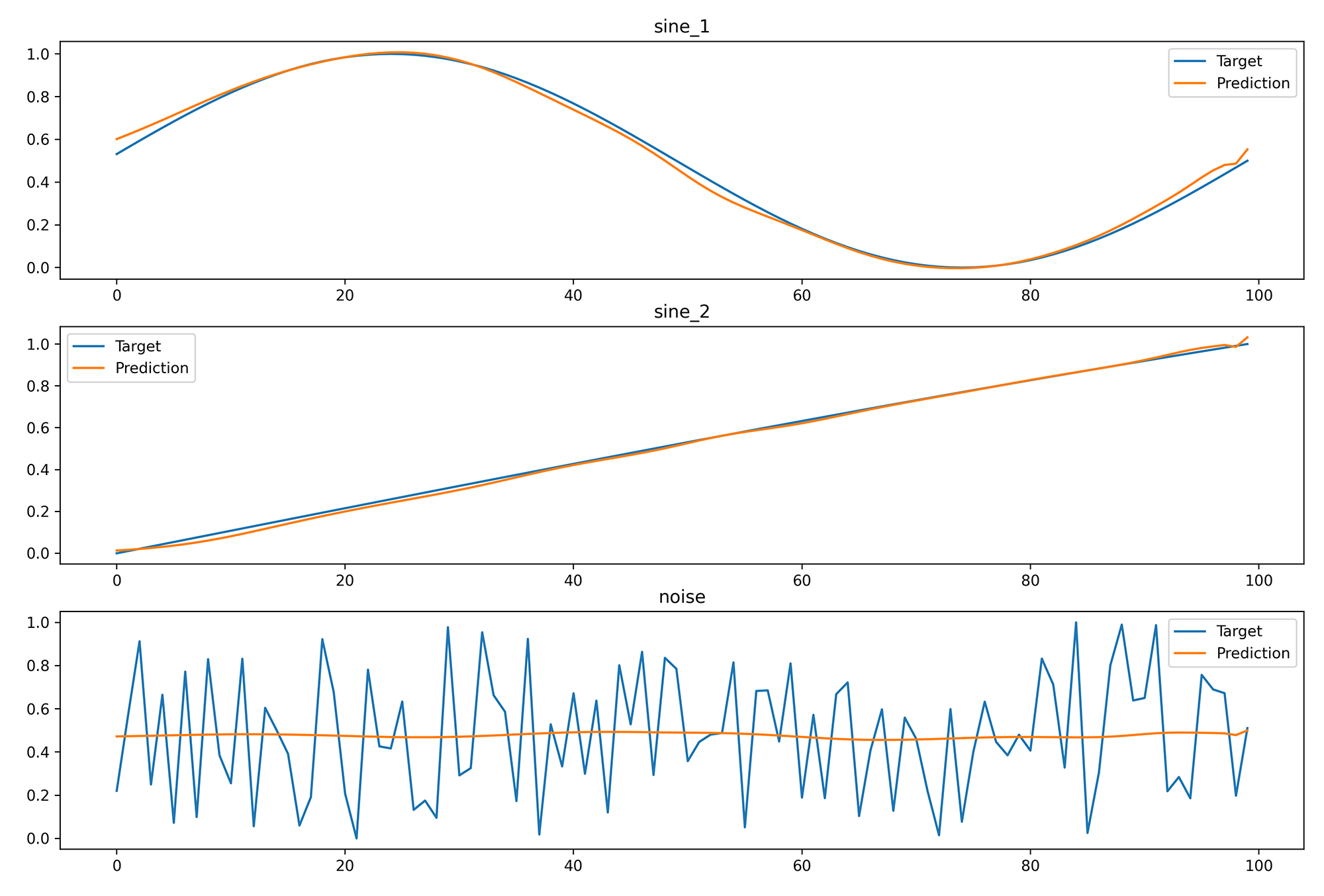}
  \caption{Visualization of the Decoder recreations of the input dimensions. Note the stochastic noise dimension has no recreatability and hence the signal makes an average prediction to minimize error}
  \label{fig:decoderrecreation}
\end{figure}

The second branch of the MvTAe model acts as a predictive branch – we call this the Alpha branch, as it generates a predictive alpha signal as its output. Its input is the output of the encoder - the hidden state vector which, when trained sufficiently, represents the underlying context and drivers of the dataset, and hence can be used to train the predictive alpha branch for a forward looking prediction of the dataset.

The structure of the alpha branch is a traditional deep fully connected one, whereby there exist two fully connected hidden layers of neurons. To allow for modelling non-linearity, which most complex sequential problems require, the activation functions of the neurons in the two hidden layers are made to be rectified linear units (ReLU). ReLU functions were chosen here as they represent the most stable functions for representing non-linearity as shown by \cite{6638312} where ReLU functions help alleviate the problem of vanishing/exploding gradients in the backpropagation process.

The target output for the alpha branch is the normalized 1-step ahead datapoint of the dimension we are looking to model for a particular data window, hence for data window $W_{(i-N) \rightarrow i)}^{k}$ we define the target as $W_{i+1}^{k}$.

As such this is the first time we use the combined signal dimension of the dataset in the model, which is ultimately the dimension we are trying to predict. It is important to note however that during the normalization process the target 1-step ahead is NOT included in the initial normalization calculation as this would lead to unwanted information leaking. As such when normalizing the target 1-step ahead datapoint we normalize this point independently with respect to the hi and lo values obtained from the respective data window normalization.

\begin{figure}[t]
  \includegraphics[width=\linewidth]{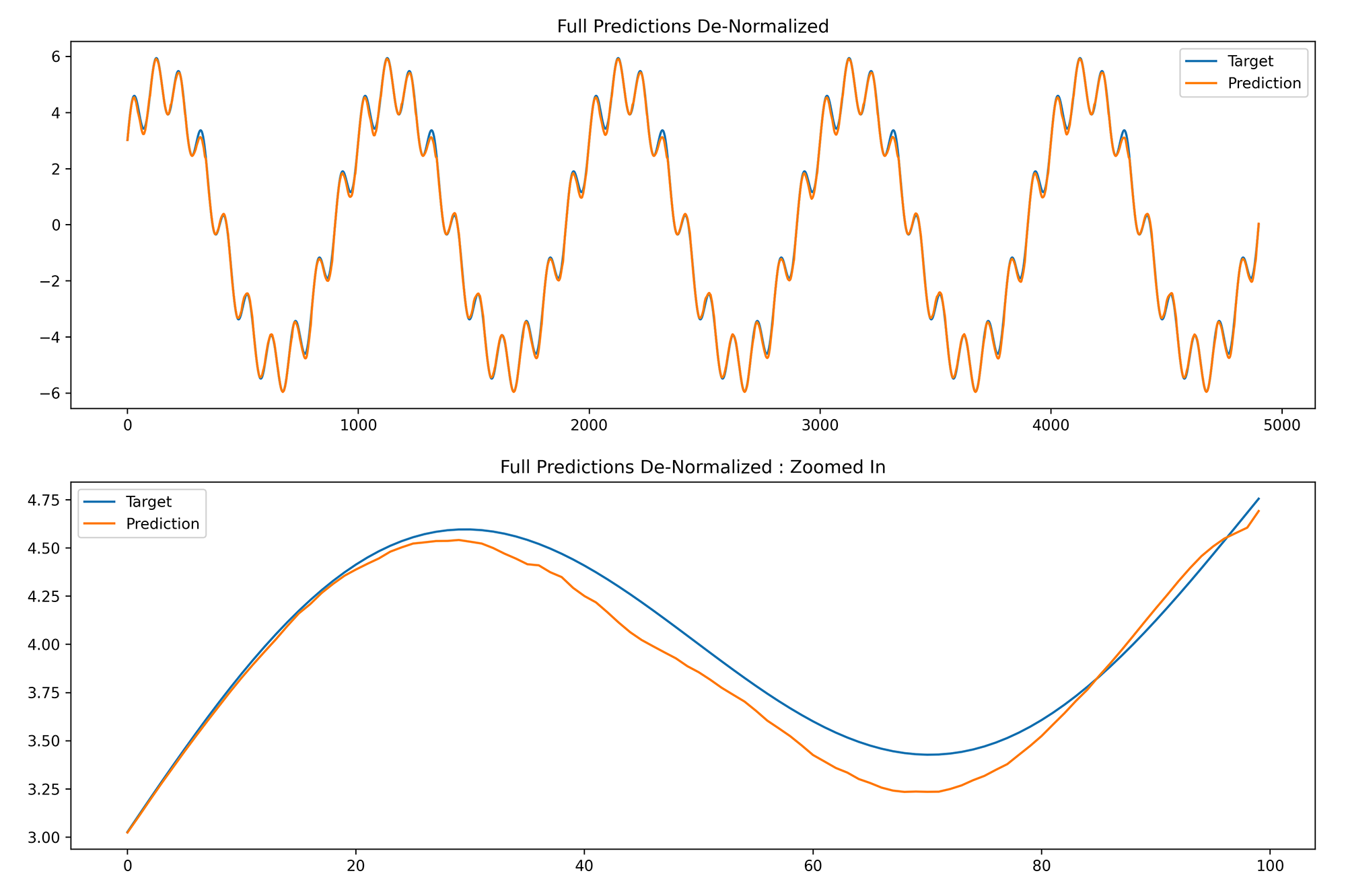}
  \caption{De-normalized predictions vs targets, de-normalized using the stored hi, lo values for each data window}
  \label{fig:denormedpreds}
\end{figure}

As with the EncoderDecoder branch, the Alpha branch is trained using the standard backpropagation algorithm \cite{134446} and with respect to a mean squared error (MSE) loss function. MSE loss is used as both problem branches (EncoderDecoder and Alpha branch) deal with regressive prediction of continuous targets rather than any classification problem. For this particular model an Adam optimizer function is employed \cite{Kingma2015AdamAM} due to the proven optimal convergence of regression problems using the Adam function.

\section{Experiments}

Result accuracy is measured using Mean Squared Error (MSE), Mean Absolute Error (MAE) and an $R^2$ value to measure the correlation between the predictions and targets. In each MSE and MAE we look to minimize the error in the first instance and maximize the $R^2$ value in the second instance by tuning the three primary drivers of our model: batch size, hidden vector size and data window size. Many other hyperparameters such as learning rate, activation function values, and neural layer sizes can also be explored, however in these experiments we only show the three drivers mentioned above which were shown to have the greatest varying influence on accuracy and the other hyperparameters are left generally optimized.

We performed a limited parameter search along the three primary hyperparameters mentioned above with a model run of 100 epochs for each search point. These parameter results can be found in Tables 1, 2, 3.

\begin{table}[!ht]
\caption{Varying batch size with a fixed hidden vector size of 64 and fixed window size of 100}
\centering
\begin{tabular}{c c c c}
\hline\hline
Batch Size & MSE & MAE & $R^2$ \\ [0.5ex]
\hline
1 & 0.00552 & 0.05372 & 96.62\% \\
2 & 0.00435 & 0.04771 & 97.34\% \\
4 & 0.00255 & 0.03621 & 98.44\% \\
8 & 0.00165 & 0.02728 & 99.01\% \\
16 & 0.00319 & 0.04205 & 98.05\% \\
32 & 0.00954 & 0.07259 & 94.16\% \\
64 & 0.02436 & 0.12341 & 85.10\% \\
128 & 0.03692 & 0.16036 & 77.43\% \\ [1ex]
\hline
\end{tabular}
\end{table}

\begin{table}[!ht]
\caption{Varying hidden vector size with a fixed batch size of 8 and fixed window size of 100}
\centering
\begin{tabular}{c c c c}
\hline\hline
Hidden Vector Size & MSE & MAE & $R^2$ \\ [0.5ex]
\hline
8 & 0.01738 & 0.09427 & 89.37\% \\
16 & 0.00940 & 0.07003 & 94.25\% \\
32 & 0.00393 & 0.04480 & 97.60\% \\
64 & 0.00165 & 0.02728 & 99.01\% \\
128 & 0.00167 & 0.02967 & 98.65\% \\
256 & 0.00323 & 0.04403 & 98.02\% \\
512 & 0.00359 & 0.04621 & 97.80\% \\
1024 & 0.00862 & 0.07046 & 94.73\% \\ [1ex]
\hline
\end{tabular}
\end{table}

\begin{table}[!ht]
\caption{Varying window size with a fixed batch size of 8 and fixed hidden vector size of 64}
\centering
\begin{tabular}{c c c c}
\hline\hline
Window Size & MSE & MAE & $R^2$ \\ [0.5ex]
\hline
5 & 0.12368 & 0.15590 & 79.58\% \\
10 & 0.04560 & 0.10294 & 88.04\% \\
25 & 0.01669 & 0.06574 & 93.80\% \\
50 & 0.01242 & 0.06452 & 93.93\% \\
100 & 0.00165 & 0.02728 & 99.01\% \\
200 & 0.00311 & 0.03918 & 98.16\% \\
400 & 0.00165 & 0.02833 & 98.99\% \\
800 & 0.00259 & 0.03608 & 97.52\% \\ [1ex]
\hline
\end{tabular}
\end{table}

\subsection{\sc{Comparison to other models}}

A standard single layer vanilla LSTM network with dropout was used as a comparison model for MvTAe. This comparison network consists of a single LSTM layer connected to a dense output layer of size one. The vanilla LSTM network was chosen for this dataset as it is best able to use the same normalized windowed data which MvTAe uses for this particular toy problem. Further classical time series forecasting models which employ autoregressive and moving average techniques - such as the popular ARIMA model – were not deemed a suitable comparison for this particular toy problem given the nature of the toy problem itself: that is the fact that it is a uniformly varying sinewave composed of several dimensions of other sinewaves. Given the autoregressive nature of a model such as ARIMA it stands to reason that it would be able to fit its coefficients perfectly to a sinewave without even the use of its regressors. This however defeats the purpose of using a windowed approach as we do with MvTAe which is more indicative of real-world problems where data sizes are usually too large and too complicated to regress over in their entirety and furthermore when real-time low latency decisions are needed usually only a window of recent data is available for processing. As such a further comparison of MvTAe vs. ARIMA (or ARMA, AR, MA style models) would be more suited to a real-world non-toy dataset.

\begin{figure}[t]
  \includegraphics[width=\linewidth]{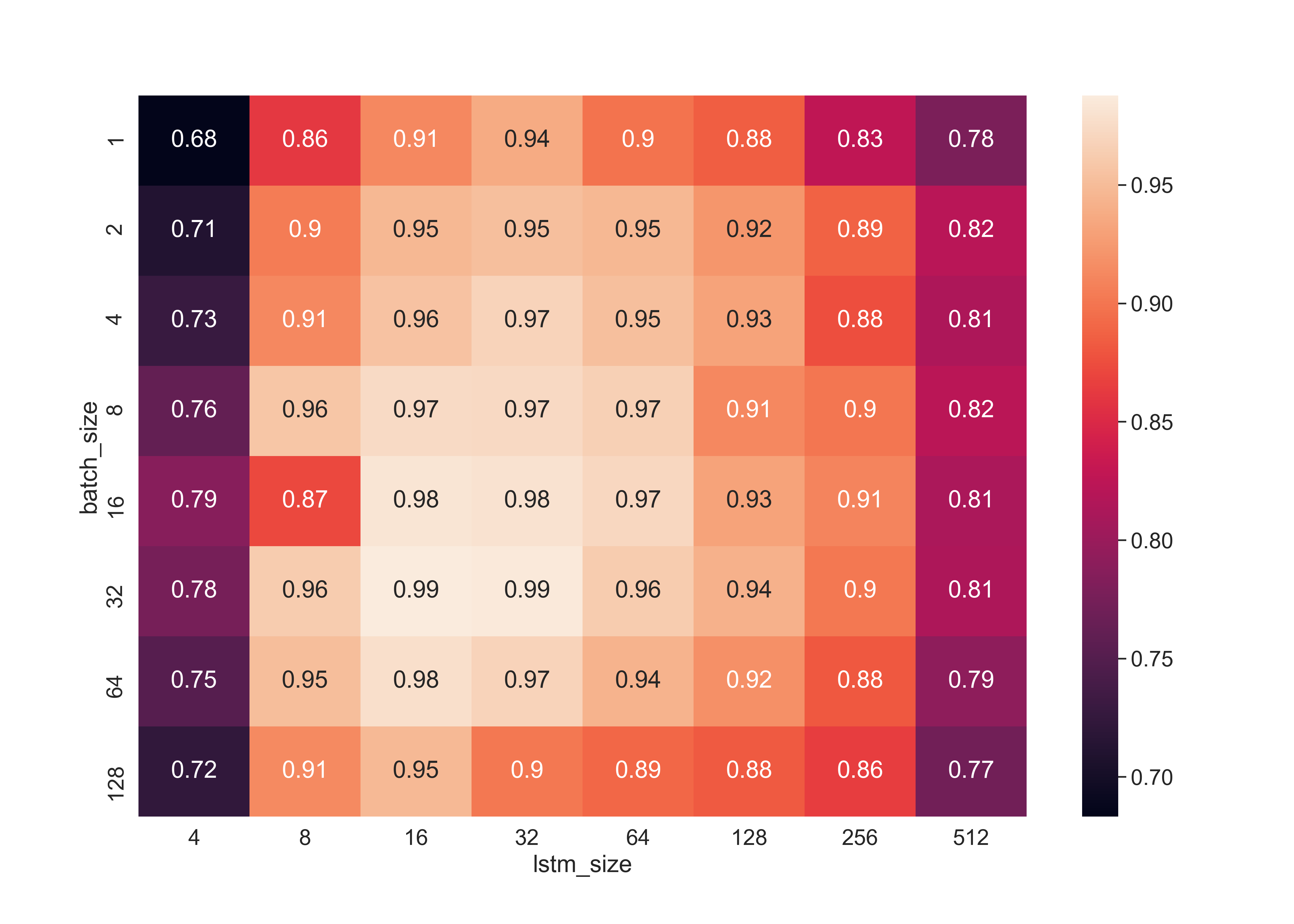}
  \caption{Heatmap of comparison model grid search results $R^{2}$}
  \label{fig:comparisonheatmap}
\end{figure}

A grid search heatmap of the vanilla LSTM models two primary drivers batch\_size and lstm\_size is shown in fig. \ref{fig:comparisonheatmap} with results as the $R^{2}$ value between predictions and true values. Note that the best result of 98.78\% falls slightly below the best result of MvTAe of 99.01\%.

\section{Conclusion}

This work shows the structure and use of a deep multi-branch neural network with a recurrent autoencoder functionality being able to successfully model a multivariate temporal data sequence by creating a hidden state vector representation of the temporal data drivers.

This is so demonstrated by using a synthetic data toy example of sine waves with various frequencies and amplitudes being combined to form a hidden target signal which the model is successfully able to recreate and forecast into the future temporal steps with excellent accuracy.

The results of the experiments with show, through a short parameter search along three primary hyperparameters of batch size, hidden vector size and data window size for 100 epochs, that the most optimal of these parameters are: batch size = 8, hidden vector size = 128, window size = 100. It is observed that there exist these optimal parameter states below which the full representation of the data cannot be captured and above which the representation is overly complex which leads to instability in accuracy.

\begin{figure}[t]
  \includegraphics[width=\linewidth]{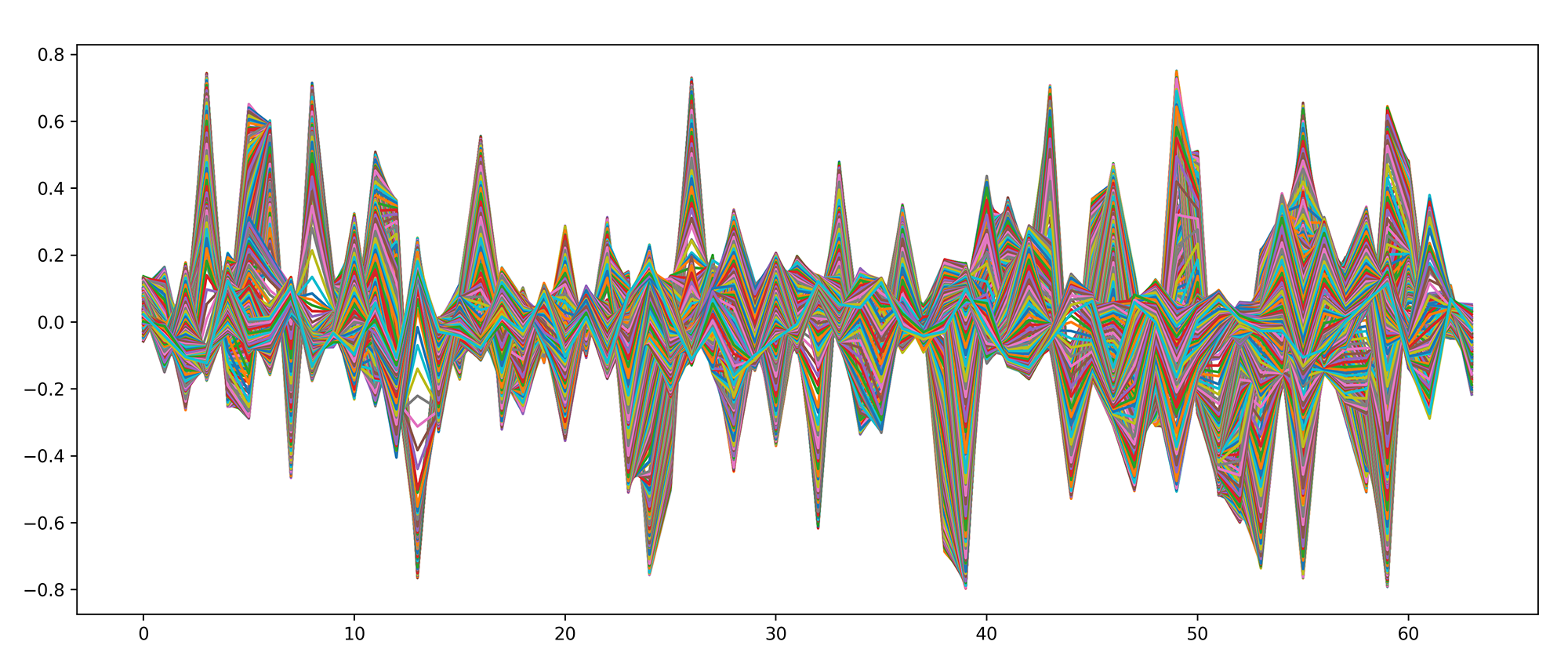}
  \caption{Visualization of the trained hidden state vector storing representations of multidimensional data sequences}
  \label{fig:hiddenvector}
\end{figure}

Interestingly it can be observed from the hidden vector size variation experiments that even with a very limited hidden vector size a reasonably accurate data window representation can be created. We see that despite the target signal being composed of 100 sequential steps of multiple dimensions, the representation of the full dimensionality of the data window can be compressed within a hidden state vector of size 8 and still retain 89.37\% accuracy.

Given the toy nature of the dataset one has to be cautious in utilizing this type of approach in the real world without some modifications. Primarily signals in real world datasets are seldom as clean as presented here and contain more noise and errors. Data collection is also a big concern with real world usage, whereby having datapoints across dimensions which are inaccurately reflected within the sequence steps can cause a model to either fail to find any meaningful state representation or, worse yet, inaccurately find a state representation from data which is leaking future information through misalignment on the temporal path.

One other point to consider is the usage of LSTM cells. Due to the nature of LSTM cells the temporal memory is reasonably short and has a tendency to decay exponentially for longer term sequences \cite{Trinh2018LearningLD}. In this model this effect is dampened through the use of inverse target sequences in the EncoderDecoder branch, however this has the negative effect of diminishing long term dependencies if they exist. Further research into using a different memory cell structure whereby long term dependencies can be more accurately captured is suggested.

Important however is the successful notion that MvTAe proves in being able to compress multivariate temporal data into single hidden vector representations and further using these static vector representations to forecast future steps in a temporal series. Perhaps with advances in measurement techniques, storage and computational power, one day we will be able to use such models to literally look steps ahead into the future of sections of the local universe, this would however have implications in a philosophical debate about the deterministic vs. stochastic nature of the universe which is a topic for a discussion orthogonal to this research.

\bibliographystyle{abbrvnat}
\bibliography{MvTAe-ResearchPaper}

\end{document}